\documentclass[letterpaper, 10 pt, conference]{ieeeconf}   

\IEEEoverridecommandlockouts                              

\usepackage{array} 
\usepackage{graphicx}
\usepackage{amsmath} 
\usepackage{comment}
\usepackage{booktabs}       
\usepackage{multirow}
\usepackage{makecell}
\usepackage{adjustbox}
\usepackage{xcolor}
\usepackage{cite}

\usepackage{bbold}
\usepackage{amsfonts}
\usepackage{amssymb}
\usepackage{txfonts}
\usepackage{pifont}

\usepackage{booktabs}
\usepackage{nicematrix}

\usepackage{algorithm,algorithmic}
\renewcommand{\algorithmiccomment}[1]

\newcolumntype{C}{>{\centering\arraybackslash}p{4.5em}}

\title{\LARGE \bf
Socratic Planner: Self-QA-Based Zero-Shot Planning for Embodied Instruction Following}

\author{Suyeon Shin\inst{1} \and
Sujin Jeon\inst{1*} \and
Junghyun Kim\inst{1*} \and
Gi-Cheon Kang\inst{1*} \and
Byoung-Tak Zhang\inst{1,2}}

\author{Suyeon Shin$^{1,2}$ $\,$ Sujin Jeon$^{1,2}$$^{*}$ $\,$ Junghyun Kim$^{1,2}$$^{*}$ $\,$ Gi-Cheon Kang$^{1,2}$$^{*}$ $\,$ Byoung-Tak Zhang$^{1,2}$
\thanks{$^*$Authors have equal contributions}
\thanks{$^{1}$AI Institute, Seoul National University}
\thanks{$^{2}$Interdisciplinary Program in AI, Seoul National University}
\thanks{\{syshin, sjjeon, junghyunkim, gckang, btzhang\}@bi.snu.ac.kr}
}

\begin{document}

\maketitle
\thispagestyle{empty}
\pagestyle{empty}

\begin{abstract}
Embodied Instruction Following (EIF) is the task of executing natural language instructions by navigating and interacting with objects in interactive environments.
A key challenge in EIF is compositional task planning, typically addressed through supervised learning or few-shot in-context learning with labeled data.
To this end, we introduce the Socratic Planner, a self-QA-based zero-shot planning method that infers an appropriate plan without any further training.
The Socratic Planner first facilitates self-questioning and answering by the Large Language Model (LLM), which in turn helps generate a sequence of subgoals.
While executing the subgoals, an embodied agent may encounter unexpected situations, such as unforeseen obstacles. 
The Socratic Planner then adjusts plans based on dense visual feedback through a visually-grounded re-planning mechanism.
Experiments demonstrate the effectiveness of the Socratic Planner, outperforming current state-of-the-art planning models on the ALFRED benchmark across all metrics, particularly excelling in long-horizon tasks that demand complex inference.
We further demonstrate its real-world applicability through deployment on a physical robot for long-horizon tasks.
\end{abstract}
\section{Introduction}
\label{sec:intro}
\begin{figure*}[t]
\centering
\includegraphics[width=15cm]{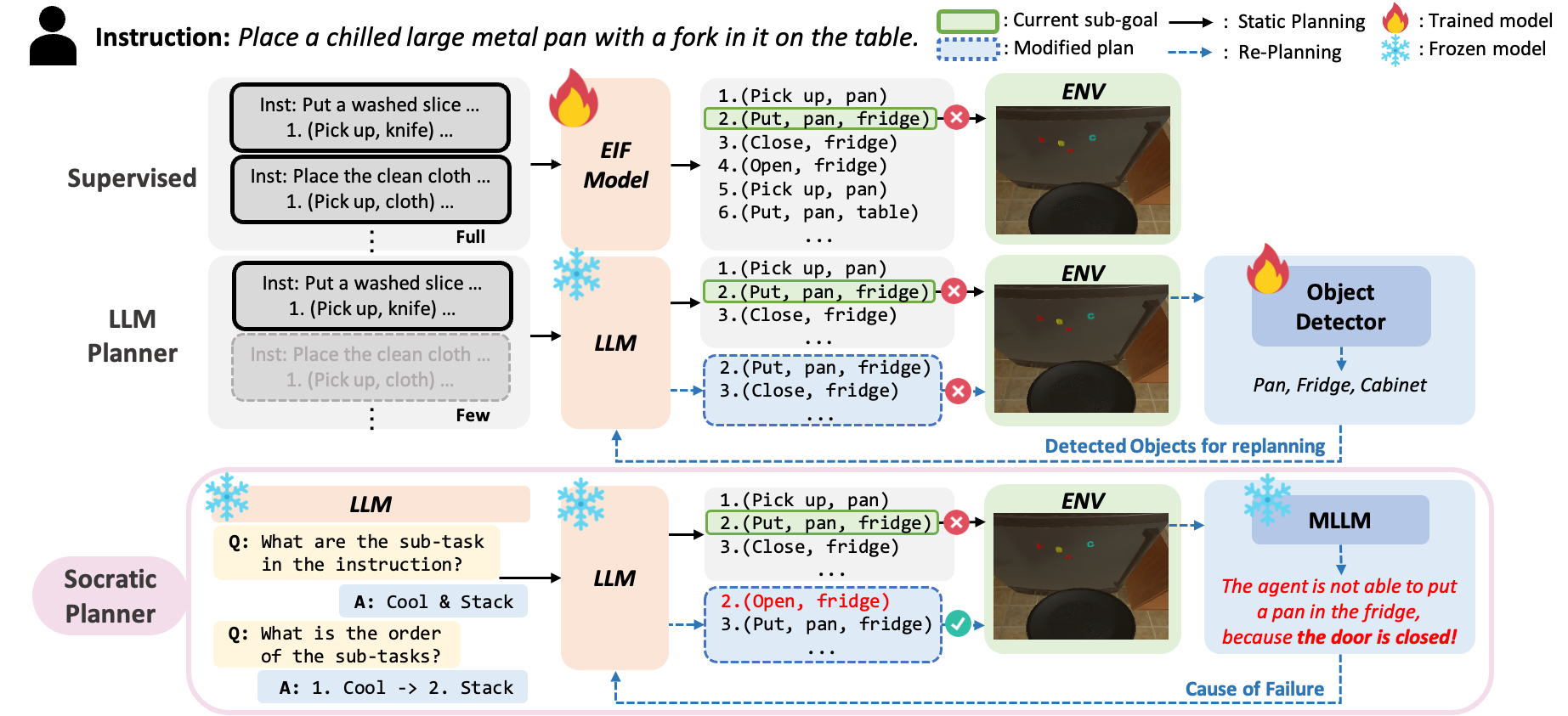}
    \caption{\textbf{Comparison between existing EIF methods and the Socratic Planner.} 
    The Socratic Planner enriches zero-shot task planning through self-questioning and answering to extract substructural information. Based on QA conversations, a Large Language Model (LLM) generates subgoals. Solid black arrows denote static planning paths, while dashed blue arrows represent re-planning paths upon encountering execution failures, with visually grounded feedback from the Multimodal Large Language Model (MLLM) guiding LLM to modify the plan.}
    \label{figure_1} 
\end{figure*}

With advancements in AI and robotics, there is a growing emphasis on developing embodied agents capable of following natural language instructions. 
Embodied Instruction Following (EIF) has emerged as a key task as it challenges agents to generate and execute a sequence of subgoals (\textit{e.g.}, \textit{(pick up, mug)}, \textit{(put, mug, sink)}, \textit{(toggle on, sink)}...), given human instruction (\textit{e.g., ``Rinse off a mug and place it in the coffee maker''}) within interactive environments.

Existing approaches in EIF primarily rely on supervised learning~\cite{ALFRED,CAPEAM,Episodic_transformer,FILM,HITUT,HLSM,LGS-RPA,Prompter,M-track,LEBP,arewethereyet,bigpicture,Look_wide,factorizing,EmbodiedBert}, where embodied agents are trained on human-annotated data consisting of instructions and corresponding expert trajectories.
However, they require large amounts of labeled data and often struggle to generalize to unseen instructions or environments.
More recent research has explored leveraging the reasoning capabilities of the Large Language Model (LLM) for task planning~\cite{LLM-Planner,progprompt,Neuro-Symbolic,jarvis,innermonologue,EmbodiedTaskPlanningwithLLM,react,thinkbot,saycan}, with most methods prompting LLM using in-context examples. 
While this paradigm significantly reduces the burden of collecting labeled data, the LLM-based planner for EIF~\cite{LLM-Planner} still necessitates a hundred human-annotated data to take advantage of in-context learning. 

In this paper, we ask: \textit{how can embodied agents perform the task of EIF without relying on any labeled data?}

Eliciting Socratic reasoning~\cite{socratic_method} from the LLM itself and further grounding it in the environment through dense visual feedback, we propose the \textbf{Socratic Planner}, a method specialized in handling complex planning tasks.
The Socratic Planner consists of three components: Socratic Task Decomposer (STD), Task Planner, and Vision-based Socratic Re-planner (VSR), as illustrated in Figure~\ref{figure_1}. 
Initially, the \textbf{Socratic Task Decomposer (STD)} uses a self-QA-based approach, leveraging LLM to decompose the instructions into substructural information needed for task execution (\textit{e.g.}, sub-tasks, order, target objects, and specific execution steps).
For example, given human instruction~\textit{``Place a chilled large metal pan with a fork in it on the table.''}, the LLM within STD engages in self-questioning and answering to decompose the task into substructural information (\textit{e.g.,} the order of the sub-tasks), aiding the Task Planner.
Based on question-and-answer conversation, the \textbf{Task Planner} synthesizes a sequence of subgoals through the LLM that can be directly applied to the controller, where each subgoal is either an action-object pair (\textit{e.g.}, (Pick up, pan)) or an action-object-receptacle triplet (\textit{e.g.}, (Put, pan, fridge)).
If the agent encounters a failure while executing a subgoal sequentially, the multimodal LLM within \textbf{Vision-based Socratic Re-planning (VSR)} provides dense visual feedback to the Task Planner through Socratic reasoning based on visual information from the environment.
For example, when the agent attempts to~\textit{(Put, pan, fridge)} in step 2 and fails, the multimodal LLM infers the cause of failure based on the current visual state of the environment. Upon inferring~\textit{``The agent is not able to put a pan in the fridge, because the door is closed''}, Task Planner successfully revises the appropriate subgoal~\textit{(Open, fridge)}. 

To demonstrate the effectiveness of the Socratic Planner, we conduct extensive experiments on the ALFRED benchmark~\cite{ALFRED}, which contains diverse environments and complex tasks for EIF. 
Our method achieves superior results compared to the state-of-the-art few-shot method~\cite{LLM-Planner} across all metrics, even in the zero-shot settings.
The Socratic Planner demonstrates substantial performance improvements, particularly in tasks with long-horizon and complex subgoal sequences, there by validating the high-dimensional zero-shot reasoning capabilities of the Socratic Planner.
Through ablation studies and qualitative evaluations, we further confirm the effectiveness of our proposed self-QA-based Socratic Task Decomposition method and Vision-based Socratic Re-planning.
Additionally, we validate the practical applicability of the Socratic Planner by deploying it on a real-world physical robot and evaluating its success rate in long-horizon tabletop manipulation tasks.

In summary, our contributions are mainly three-fold. 
(1) We propose the Socratic Planner, which enhances the zero-shot task planning capabilities of the Large Language Model (LLM) through two key mechanisms: Self-QA-based Socratic Task Decomposition (STD) and Vision-based Socratic Re-planning (VSR).
(2) We extensively validate the performance of the Socratic Planner using a 3D simulator and further demonstrate its practical applicability in real-world settings by deploying it on a physical robot.
(3) Experimental results demonstrate that our approach effectively performs complex reasoning for EIF without the need for labeled examples and outperforms state-of-the-art few-shot methods.

\section{Related work}

\subsection{Embodied Instruction Following (EIF)}
EIF is the task of executing natural language instructions, where embodied agents should navigate and interact with objects in environments~\cite{ALFRED,iqa}.
Previous approaches in EIF~\cite{ALFRED,bigpicture,hierarchical_attention_model,Look_wide,Episodic_transformer,M-track,EmbodiedBert,HITUT,arewethereyet} rely on supervised learning, which requires extensive human-annotated data and often struggles to generalize to novel environments.
Meanwhile, recent studies leverage the Large Language Model (LLM) for planning~\cite{LLM-Planner,thinkbot,languagemodelaszeroshot,innermonologue,Neuro-Symbolic}, with some using LLM as auxiliary helpers~\cite{thinkbot}. 
Several methods~\cite{languagemodelaszeroshot,LLM-Planner,innermonologue,Neuro-Symbolic} use in-context examples as text prompts to guide the LLM in generating plans.
Our approach belongs to this paradigm. However, instead of injecting in-context samples into LLM, the Socratic Planner is the first to address EIF through zero-shot prompting without any labeled data.

\subsection{Multi-Step Reasoning with Large Language Model (LLM)}
Our work is closely related to the large body of research on multi-step reasoning that uses LLM to decompose complex problems into intermediate steps.
Chain-of-thought (CoT) prompting~\cite{wei2022chain} is one of the most representative methods prompting LLM to generate intermediate reasoning steps explicitly.
Notably, a line of research~\cite{zhou2022least,shridhar2022distilling,press2022measuring} has focused on decomposing a complex problem into more manageable sub-problems to enhance the problem-solving capabilities of LLM, especially given a few shot samples containing subquestion-solution pairs.
Our approach follows the latter line of work but addresses the inherent complexity of EIF, which requires long-horizon task planning and involves diverse scenarios.
Therefore, with only a few shot samples, enabling the LLM to perform self-QA for EIF is challenging. 
Moreover, manually creating such pairs is inefficient and time-consuming.
Instead of injecting in-context samples, we define key elements to be discovered, which could be generally applicable in EIF tasks, allowing LLM to autonomously generate and answer questions via zero-shot prompting.

\subsection{Re-planning}
Agents executing generated plans often encounter failures due to various factors like controller uncertainty or unexpected environment states. 
Thus, iterative plan adjustments are crucial. 
However, most EIF studies~\cite{ALFRED,arewethereyet,bigpicture,CAPEAM,EmbodiedBert,Episodic_transformer,FILM,HLSM,LEBP,LGS-RPA,M-track,saycan,thinkbot,Prompter,languagemodelaszeroshot} overlook re-planning mechanisms. 
A work~\cite{HITUT} introduced a simple heuristics-based re-planning method, where the agent backtracks and retries the previous step upon subgoal failure.
However, this approach is limited to addressing failures caused by incorrect predictions.
Recent research~\cite{progprompt,choi2024lota,innermonologue,reprompting,LLM-Planner} explore agent-driven plan modifications using online feedback. 
Some studies~\cite{progprompt,choi2024lota,innermonologue} provide feedback to the agent through human input or simulation message, which is impractical in real-world scenarios since error messages may not be defined or human feedback may be unavailable. 
Another approach, named re-prompting~\cite{reprompting}, uses predefined errors and template-based prompts but lacks visual information, limiting its ability to handle unpredictable failures. 
LLM-Planner~\cite{LLM-Planner} updates plans based on visible object lists but often fails as it relies solely on the object lists, lacking a comprehensive understanding of the current visual state, as illustrated in Figure~\ref{figure_1}.
To address the aforementioned limitations, we propose Vision-based Socratic Re-planning (VSR), which integrates multimodal LLM~\cite{gpt4o} to provide visually grounded feedback. 
VSR enhances plan adjustments by using detailed visual information (\textit{e.g.}, scene depiction, subgoal feasibility, and failure causes), maximizing the LLM's reasoning abilities without relying on human or system message.
\section{Method}

\begin{figure}[!t]
\centerline{\includegraphics[width=\columnwidth]{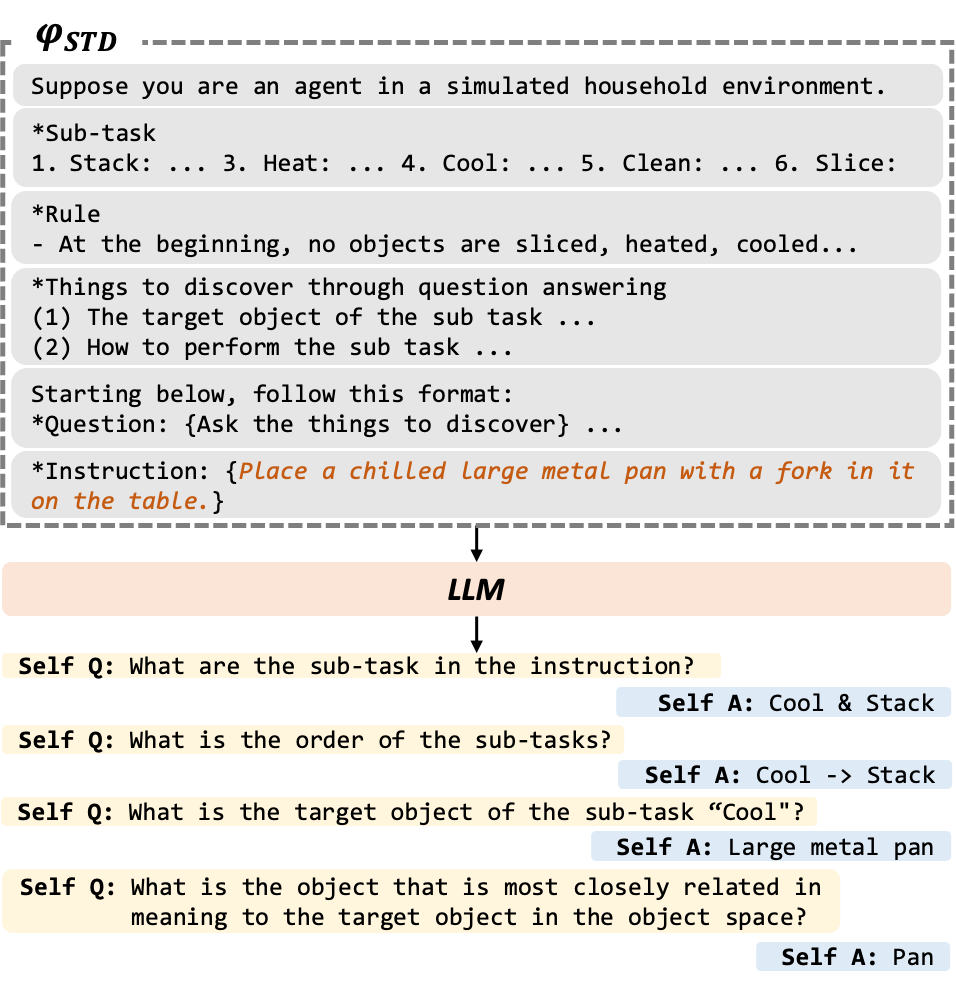}}
    \caption{\textbf{Input prompt and output of the Socratic Task Decomposer.}}

    \label{fig:figure_std} 
\end{figure}

\subsection{Task Definition}
EIF executes a sequence of actions implied in a natural language instruction $I$ in interactive environments. 
This task is divided into two stages: subgoal planning and low-level control. First, subgoal planning predicts a sequence of subgoals. 
For instance, a plan for the instruction \textit{``put a heated slice of bread in the fridge''} can be translated into a series of subgoals: [(pickup, knife), (slice, bread), ..., (close, fridge)].
Each subgoal consists of (1) an admissible action (\textit{e.g.}, Pickup, Put, ToggleOn, ToggleOff, Open, Close, Slice), (2) a target object, and optionally, (3) a receptacle (\textit{i.e.}, target place).
Then, the low-level controller translates each subgoal into primitive navigation actions interacting with environments.
Accordingly, our focus lies in subgoal planning, which necessitates complex, multi-step reasoning. We integrate the Socratic Planner with the low-level controller~\cite{HLSM}, same as the baseline models~\cite{LLM-Planner}.

\begin{algorithm}
\caption{Socratic Planner}
\label{alg:socratic_planner}
\begin{algorithmic}[1]
\small
\REQUIRE $I \gets$ Human Instruction
\REQUIRE Controller, Object-Detector, LLM, MLLM
\REQUIRE Prompt generators: $\varphi_{*}$ (\ref{sec:decomposing},\ref{sec:policy},\ref{sec:Re-planning})
\REQUIRE Set of Observed Objects, $\mathcal{O} \gets \emptyset$
\STATE \textbf{Define} subgoal $\triangleq (\text{action}_{sg}, \text{object}_{sg}, \text{receptacle}_{sg})$
\STATE \textbf{Define} $\mathcal{P}$ as list of subgoals
\STATE \textbf{Define} \textit{scene} $\gets$ a RGB image of current camera output
\STATE $\mathcal{QA}\gets$ LLM$(\varphi_{STD}(I))$ \COMMENT{~\cref{sec:decomposing}}
\STATE $\mathcal{P} \gets $ LLM$(\varphi_{TP}(I,\mathcal{QA}))$ \COMMENT{~\cref{sec:policy}}
\FOR{subgoal \textbf{in} $\mathcal{P}$} 
    \STATE execute Controller$(\text{subgoal})$, updating \textit{scene} \COMMENT{~\cref{sec:Re-planning}}
    \STATE $\mathcal{O} \gets \mathcal{O}\, \cup$ Object-Detector$(scene)$
    \IF{execution \textit{fails}}
        \STATE $v \gets$ MLLM$(\varphi_{validity}(\text{subgoal}),scene)$
        \IF{$\text{object}_{sg}$ not in $\mathcal{O}$ \textbf{or} $v$ is $invalid$} %
            \STATE $f \gets$MLLM$(\varphi_{VSR}(\text{subgoal}, v),scene)$
            \STATE $\mathcal{P} \gets$LLM$(\varphi_{TP}(f,\mathcal{P}, \mathcal{O},v))$
        \ELSE
            \STATE $\text{redo}$ \COMMENT{Redo the current subgoal}
        \ENDIF 
    \ENDIF
\ENDFOR
\end{algorithmic}
\end{algorithm}

\subsection{Overview}
The research community has observed the robust zero-shot reasoning abilities of the Large Language Model (LLM)~\cite{gpt,gpt4,gpt4o} across various tasks. 
However, zero-shot inference on complex tasks like EIF remains challenging due to the need for high-level reasoning. 
To tackle this, our Socratic Planner employs self-questioning and answering for complex reasoning through three components.
Through self-QA, \textit{The Socratic Task Decomposer} (STD) first decomposes tasks into the substructural information required to complete the task (\ref{sec:decomposing}).
Based on the self-QA conversation, \textit{Task Planner} synthesizes a subgoal sequence that aligns with the low-level controller's input format (\ref{sec:policy}).
Next, the agent executes these subgoals sequentially.
If execution fails, the Socratic Planner iteratively adjusts the plan using dense visual feedback, a process called \textit{Vision-based Socratic Re-planning} (VSR, \ref{sec:Re-planning}).
GPT-4o~\cite{gpt4o} is used throughout. 

\subsection{Socratic Task Decomposer}
\label{sec:decomposing}
Natural language instructions in EIF are inherently compositional, making it natural to break them down into more fine-grained elements.
Inspired by Socrates's Socratic method~\cite{socratic_method}, a method used to stimulate critical thinking through a series of questions and answers, we empower LLM to decompose complex task instructions into substructural information via self-questioning and answering. 
This process, called the Socratic Task Decomposer (STD), enables the high-dimensional inference required in EIF.
Before describing the details of STD, the ``sub-task'' refers to the intermediate-level task. 
For example, the instruction ~\textit{``put a heated slice of bread in the fridge''} includes three sub-tasks~\textit{``heating the bread''},~\textit{``slicing bread''}, and~\textit{``placing bread in the fridge''}. Each sub-task consists of multiple subgoals (\textit{e.g., ``heating the bread''}: [(Pickup, bread), (Navigate, microwave), (Open, microwave)...]). 
The depth and content of substructural information depend on task complexity. 
To aid LLM in autonomous reasoning about the complex instruction for general EIF tasks, we provide guidance in the form of ``things to discover'', prompting LLM to consider essential aspects such as identifying constituent sub-tasks, sequencing, target objects, and specific task execution methods. 
As depicted in Figure~\ref{fig:figure_std}, the prompt for STD comprises the aforementioned ``things to discover'', along with the role of the LLM, sub-task explanations, and simulator rules. 
The LLM then generates question-and-answer sequences ($\mathcal{QA}$) to discover the substructural information for task completion.

\subsection{Subgoal Generation}
\label{sec:policy}
Utilizing the $\mathcal{QA}$ generated in~\ref{sec:decomposing}, the Task Planner generates a subgoal sequence that aligns with the input format of the low-level controller. 
The prompt generator for Task Planner, denoted as $\varphi_{TP}$, takes $I$ and $\mathcal{QA}$ as input to produce a prompt.
$\varphi_{TP}$ is augmented with simulator rules and output format, along with the command to create a detailed plan (\textit{i.e., ``Based on this conversation, create a detailed plan for executing instructions that consist of various sub-tasks''}).
Subsequently, this prompt is fed into the LLM to generate the sequence of subgoals, where each subgoal is a combination of primitive actions, interacting objects, and, optionally, receptacles. $\mathcal{P} \gets \text{LLM}(\varphi_{TP}(I,\mathcal{QA}))$.

\subsection{Vision-based Socratic Re-planning}
\label{sec:Re-planning}
Vision-based Socratic Re-planning (VSR) enables agents to recover from execution failures by adjusting the initial plan through dense visual feedback to better suit the physical environment.
Starting from the $6^{th}$ line of Algorithm~\ref{alg:socratic_planner}, the sequence of subgoals generated in~\ref{sec:policy} is executed sequentially by the low-level controller. 
As execution proceeds, the agent continuously updates its observed scene and adds detected objects from the controller to the set of observed objects.
When execution encounters failures, the agent seeks guidance from the Multimodal LLM (MLLM)~\cite{gpt4o}. 
The re-planning process starts with the agent determining whether the current subgoal can be executed in the environment (\textit{i.e.,} validity, denoted as $v$), and autonomously decides whether the re-planning is required. 
If the current subgoal's target object is visible and the action seems valid, the agent assumes that the failure is due to a controller malfunction and retries the subgoal again.
Otherwise, the MLLM generates feedback through QA (\textit{i.e.,} a cause of the failure, specific corrective actions), denoted as $f$, based on the cues obtained from the MLLM itself.
Leveraging QA-based dense visual feedback, the Task Planner revises the plan to fit the environment better. 
Execution then resumes from the revised subgoal, and the low-level controller continues the action sequence. 
This iterative process, called Vision-based Socratic Re-planning (VSR), grounds the LLM's reasoning in the environment.

\section{Experiment}
\begin{table*}[tb]
\caption{
\begin{flushleft}
\textnormal{
\textbf{Comparison with the state-of-the-art methods on SR and GC across four splits.}  Static models are without the visually grounded re-planning, denoted as \textdagger\ . Bold symbols in numbers denote the highest accuracy.}
\end{flushleft}
}
\label{tab:1}

\centering
\begin{tabular}{@{}lcccccccccccccccc@{}}
\toprule
\multirow{2}{*}{Model} & \multirow{2}{*} & \multicolumn{2}{c}{Valid Seen} & \multicolumn{2}{c}{Valid Unseen} & \multicolumn{2}{c}{Test Seen} & \multicolumn{2}{c}{Test Unseen} \\ \cmidrule(l){3-4}  \cmidrule(l){5-6} 
 \cmidrule(l){7-8} \cmidrule(l){9-10}
&n-shot&SR&GC&SR&GC&SR&GC&SR&GC  \\ \midrule
HLSM~\cite{HLSM}&full& 29.63& 38.74& 18.28& 31.24& 25.11& 35.79& 20.27& 27.24\\
FILM~\cite{FILM}&full& -& -& -& -& 25.77& 36.15& 24.46& 34.75\\
CAPEAM~\cite{CAPEAM}&full& -& -& -& -& 47.36& 54.38& 43.69& 54.66\\ \midrule
LLM-Planner\textsuperscript{\dag}~\cite{LLM-Planner}&few& 11.82& 23.54& 11.10& 22.44& 13.05& 20.58& 11.58& 18.47 \\
LLM-Planner~\cite{LLM-Planner}&few& 13.53& 28.28& 12.92& 25.35& 15.33& 24.57& 13.41&22.89 \\ \midrule
LLM-Planner\textsuperscript{\dag}~\cite{LLM-Planner} &zero& 5.73& 11.24& 4.39& 12.97& 5.71& 10.42& 6.54& 12.12 \\
Socratic-Planner\textsuperscript{\dag}~\textrm{(Ours)}&zero& 13.41& 27.61& 14.62& 28.60& 18.59& 30.10& 12.75& 24.54 \\
Socratic-Planner~(Ours)&zero& \textbf{18.78}& \textbf{32.20}& \textbf{15.35}& \textbf{29.20}&\textbf{18.66} & \textbf{30.15}& \textbf{13.47}& \textbf{24.89}\\ \bottomrule
\end{tabular}
\end{table*}

\subsection{Dataset}
We validate Embodied Instruction Following (EIF) capabilities of Socratic Planner using the ALFRED~\cite{ALFRED}, a representative benchmark in EIF known for its complex scenarios and diverse range of objects.
This benchmark evaluates a mapping from natural language instructions and egocentric vision to sequences of actions for long-horizon household tasks. The ALFRED dataset consists of 25k language instructions describing 8k expert demonstrations, including seven admissible actions, 108 distinct objects, and seven types of tasks in 120 scenes. 
\subsection{Evaluation Metric}
We use standard evaluation metrics in the ALFRED benchmark, Success Rate (SR) and Goal Conditioned (GC) success rate of tasks. SR represents the percentage of tasks wherein every environmental state after each action step satisfies predefined goal conditions. \textit{i.e.}, a task is considered a success only when all subgoals are achieved. On the other hand, GC measures the percentage of completed goal-conditions for each task. 
We also assess subgoal planning accuracy (Subgoal ACC), which measures how accurately the subgoals are predicted, excluding the potential malfunctions of the low-level controller.

\begin{figure}[tb]
  \centering
  \includegraphics[height=4cm]{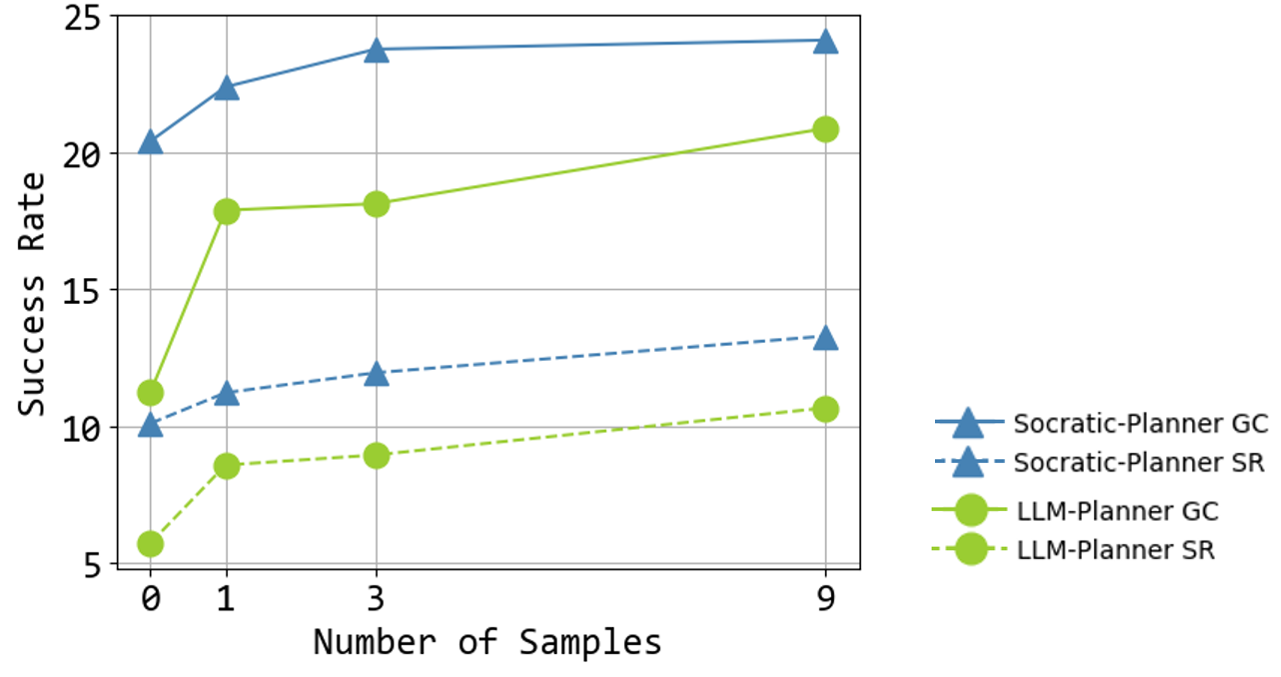}
      \caption{\textbf{Comparison with LLM-Planner on zero-shot and few-shot settings in EIF metrics}, where valid seen splits are used for evaluation.}
    \label{fig:figure_3}
\end{figure}

\subsection{Comparison with SOTA}
\label{sec:result}
\noindent \textbf{Compared Models.}
The compared models are categorized into three groups, each with its distinct training setup: fully-supervised setting~\cite{HLSM,FILM,CAPEAM}, few-shot setting~\cite{LLM-Planner}, and zero-shot setting. 
In the fully-supervised setting, models are trained using the entire training dataset. 
For the few-shot setting, the model engages in in-context learning by prompting a few samples extracted from the training data. 
Finally, the zero-shot setting entails models inferring high-level plans without utilizing any training data.

\noindent \textbf{Results on Embodied Instruction Following (EIF).}
As shown in Table~\ref{tab:1}, Socratic Planner significantly outperforms LLM-Planner~\cite{LLM-Planner} in both the zero-shot setting and the few-shot setting, where the latter is trained with nine of the most similar instruction-trajectory samples. We also observe that the use of vision-based Socratic re-planning (VSR) consistently boosts SR and GC compared with the Socratic Planner without VSR (\textit{i.e.,} Static).
Moreover, we visualize the performance in $n$-shot settings, $n \in \left\{0, 1, 3, 9\right\}$.
For n-shot comparison, GPT-3.5-Turbo~\cite{gpt} was used due to cost constraints.
As depicted in Figure~\ref{fig:figure_3}, the Socratic Planner outperforms all settings compared with LLM-Planner on both SR and GC.

\noindent \textbf{Results on Subgoal Planning (Subgoal ACC).} 
We conduct a task-type analysis to identify the accuracy of subgoal planning for each task in the ALFRED benchmark. 
In this analysis, we compare the Socratic Planner, which uses zero-shot reasoning, with the original LLM-Planner, which relies on few-shot reasoning.
In Table~\ref{tab:2}, the Socratic Planner shows improved subgoal planning accuracy in six out of seven tasks. It is also noteworthy that the Socratic Planner yields strong Subgoal ACC performance, especially in long-horizon tasks such as Heat, Cool, and Clean.
This showcases that the reasoning capabilities of the Socratic Planner are relatively robust on challenging EIF tasks.   
Meanwhile, the performance of the Socratic Planner appears to be relatively lower in the ``Pick Two'' and ``Stack'' tasks. Both tasks involve picking up two objects and relocating them. 
However, the AI2-THOR~\cite{ai2thor} simulator for the ALFRED benchmark does not support picking up two objects simultaneously. 
We believe that this unique constraint is challenging, especially for the zero-shot methods that do not observe any prior data.   

\begin{table}[tb]
\caption{
\begin{flushleft}
\textnormal{
\textbf{Subgoal ACC for LLM-Planner and Socratic Planner by task type in the static planning setting}, using valid unseen splits.}
\end{flushleft}
}
\label{tab:2}
\centering
\scalebox{1}{
\begin{tabular}{@{}lccccc@{}}
\toprule
Task  & \makecell{Task } & \makecell{LLM} & \makecell{Socratic} \\ 
Type & Length & Planner & Planner \\
\midrule
Heat  &12.1 & 16.18 & \textbf{39.71} \scriptsize{(+23.53)} \\
Cool  &9.7 & 10.09 & \textbf{68.81} \scriptsize{(+58.72)}\\ 
Clean &7.5 & 16.81 & \textbf{78.76} \scriptsize{(+61.95)} \\
Pick Two &7.3 &12.35 & \textbf{16.05} \scriptsize{(+3.70)}\\
Stack &5.0 & \textbf{11.01} & \textbf{11.01} \scriptsize{(+0.00)}\\
Pick  &3.4 & 16.00 & \textbf{43.00} \scriptsize{(+27.00)} \\
Examine &2.3 & 36.99 & \textbf{40.46} \scriptsize{(+3.47)}\\ \bottomrule
\end{tabular}
}
\end{table}

\begin{table}[tb]
\centering
\caption{
\begin{flushleft}
\textnormal{
\textbf{Ablation of Socratic Planner's STD in the static setting,} where valid unseen splits are used for evaluation.}
\end{flushleft}
}
\label{tab:3}
\scalebox{1}{
\begin{tabular}{@{}lccccccc@{}}
\toprule[1.5pt]
\multirow{2}{*}{Model} & \multicolumn{3}{c}{Valid Unseen}  \\ \cmidrule(l){2-4} 
 & Subgoal ACC & SR & GC   \\ \midrule
Socratic-Planner w.o. STD &  25.94 &9.00 & 22.20 \\
Socratic-Planner w. CoT~\cite{kojima2022large} & 23.39& 7.67&20.61 \\ \midrule
Socratic-Planner & \textbf{43.36}& \textbf{14.62}& \textbf{28.60} \\
\bottomrule[1.5pt]
\end{tabular}}
\end{table}

\subsection{Ablation Study}
\label{sec:ablation}
To demonstrate the effectiveness of individual components of the Socratic Planner, we perform an ablation study comparing models with and without Socratic Task Decomposer (STD). 
In Table~\ref{tab:3}, we can clearly identify the benefits of the STD across all evaluation metrics. 
It indicates that discovering substructural information -- represented as explicit self question-and-answer pairs helpful for synthesizing a sequence of subgoals -- is crucial for EIF. 
However, a question still remains as to why the information should be a form of question-and-answer. 
Therefore, we compare the Socratic Planner with one of the most representative multi-step reasoning methods for LLM, Chain-of-though (CoT) prompting~\cite{kojima2022large}. 
Specifically, we reproduce zero-shot CoT~\cite{wei2022chain} by replacing only the description about questioning and answering in the prompts $\varphi_{STD}(I)$ with the sentence, ``How can you decompose the instruction: $\left\{\text{instruction}\right\}$? Let's think step by step.'' 
In Table~\ref{tab:3}, the zero-shot CoT struggles to yield useful information. 
In our manual inspection, this method extracts information in a haphazard manner, which ultimately hinders the ability to handle the complex reasoning required for EIF tasks.

\begin{table}[tb]
\centering
\caption{
\begin{flushleft}
\textnormal{\textbf{
Results of Real-World Tabletop Manipulation Experiment.}}
\end{flushleft}
}
\label{tab:4}
\scalebox{1}{
\begin{tabular}{@{}lccccccc@{}}
\toprule[1.5pt]
Model &n-shot& Subgoal ACC & SR   \\ \midrule
LLM-Planner~\cite{LLM-Planner} &9&  61.7 &46.7 \\
Socratic-Planner w.o. STD &0& 48.3&38.3 \\ \midrule
Socratic-Planner &0& \textbf{81.7}& \textbf{53.3} \\
\bottomrule[1.5pt]
\end{tabular}}
\end{table}

\subsection{Real-World Experiment}
\begin{figure}[!t]
\centerline{\includegraphics[width=\columnwidth]{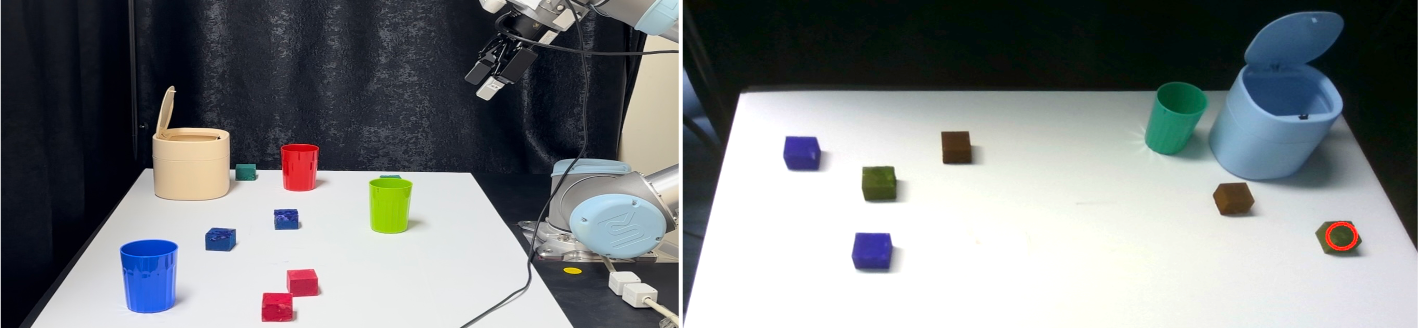}}
    \caption{\textbf{Experimental setup (left) and Robot's camera view (right)}}

    \label{experiment_setup} 
\end{figure}

\noindent \textbf{Robot Platform.}
As shown in Figure~\ref{experiment_setup}, we perform experiments with a physical robot arm, specifically the 6-DoF UR5e equipped with a two-fingered gripper. Our system is complemented by the Intel Realsense Depth Camera D435 to get an RGB-D image. The remote server handles all inferences for the Socratic Planner, and manipulation planning is performed locally on the platform.

\noindent \textbf{Evaluation Protocol.}
To verify applicability in real-world scenarios, we evaluate the Socratic Planner's ability for long-horizon planning in tabletop manipulation. We conducted an experiment with 14 objects set on a table and five possible actions: (Pickup, Put, Open, Close, Push), creating a total of 60 long-horizon instructions such as \textit{``Arrange the red and blue blocks in their corresponding boxes, then stack the cups in reverse color box."}, each requiring an average of 8.8 sub-goals to complete. We compare Socratic Planner with LLM-Planner~\cite{LLM-Planner} and Socratic Planner without STD. Humans assess both success rate and subgoal planning accuracy.

\noindent \textbf{Results.}
In Table~\ref{tab:4}, the Socratic Planner achieves a 53.3\% overall execution success rate and 81.7\% subgoal accuracy, outperforming all baselines.
Applying STD boosted performance, increasing Subgoal ACC by 33.4\% and SR by 15\%, highlighting the effectiveness of extracting key information through self-QA for structured planning. 
Unlike the LLM planner, which generates plans by imitating nine instruction-trajectory samples through in-context learning, the Socratic Planner demonstrates the advantage of complex reasoning through structured inference without additional human labor for annotation. 
This experiment also showcases the Socratic Planner’s zero-shot reasoning capability and its potential to generalize to real-world tasks.

\subsection{Qualitative Results}
\begin{figure}[!t]
\centerline{\includegraphics[width=\columnwidth]{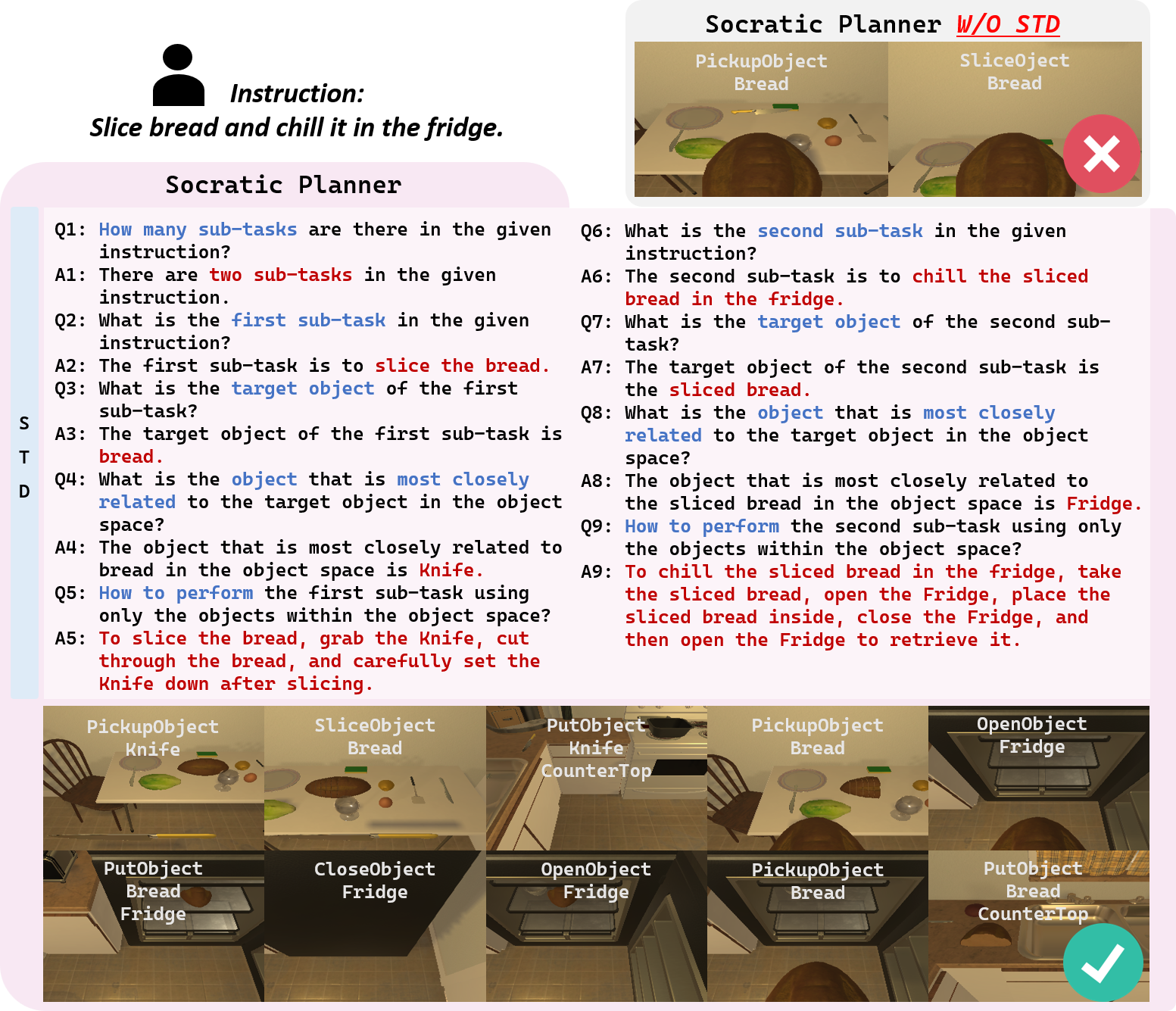}}
    \caption{\textbf{Visualization of the agent action sequence acquired by Socratic Planner without STD (top right) and our Socratic Planner (bottom)}.}
    \label{figure_std} 
\end{figure}

To visualize the effects of STD, Figure~\ref{figure_std} shows the action sequence generated by the Socratic Planner, with and without STD. 
Without STD, the agent struggles to follow the instruction \textit{``Slice bread and chill it in the fridge''} due to the lack of structured information needed to complete the instruction, leading to failure -- attempting to slice bread without first picking up a knife. 
In contrast, the Socratic Planner effectively decomposes the task into sub-tasks, the sequence of sub-tasks, and the necessary objects through self-QA. 
Based on this structured information, the agent accurately predicts and successfully executes the lengthy sequence of subgoals, demonstrating the power of structured information in task planning.

\begin{figure}[!t]
\centerline{\includegraphics[width=\columnwidth]{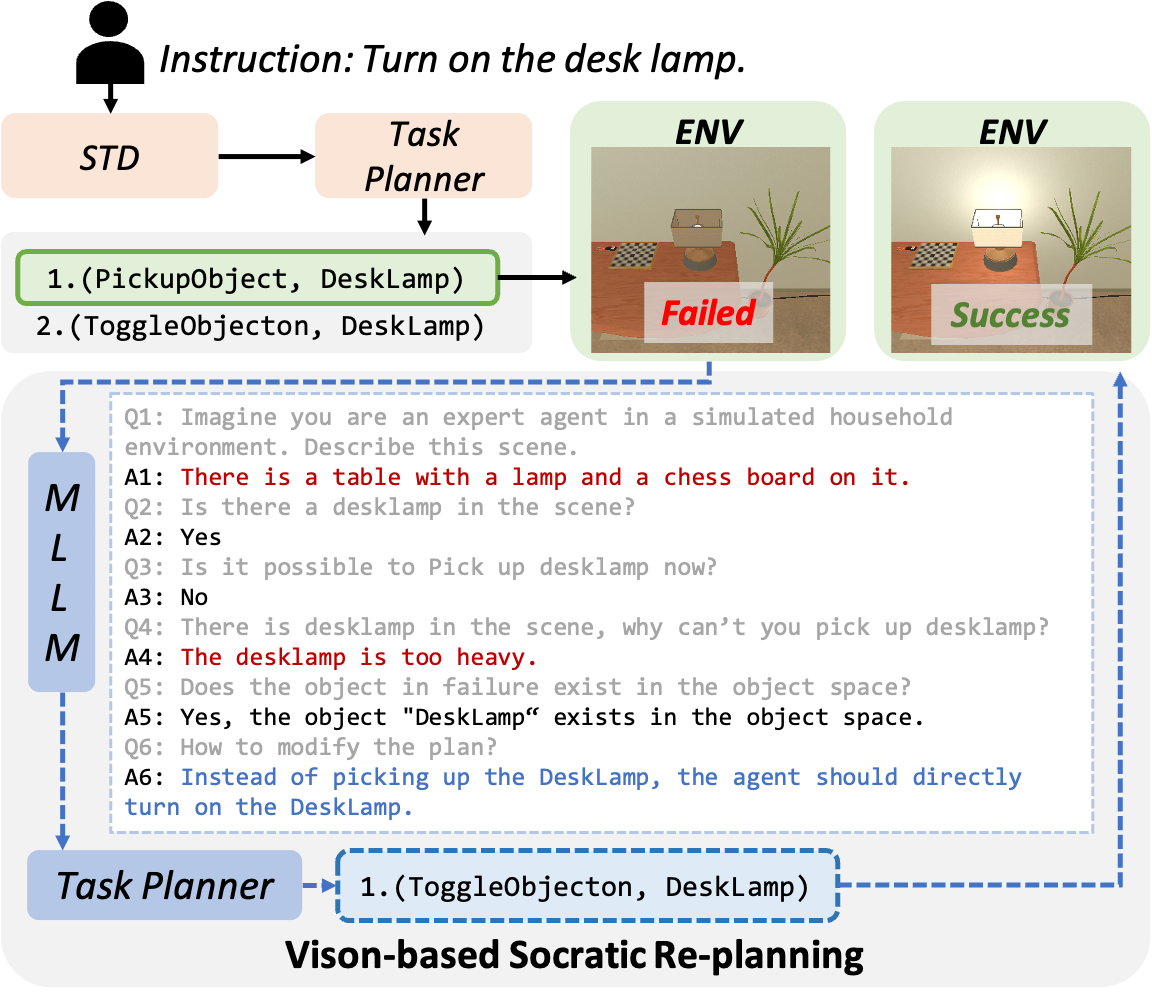}}
    \caption{\textbf{Illustration of Vision-based Socratic Re-planning.}}
    \label{figure_replan} 
\end{figure}

Figure~\ref{figure_replan} shows vision-based Socratic Re-planning. When asked to \textit{``turn on the desk lamp''}, the static Socratic Planner fails by trying to pick up the lamp, which is impossible. 
Using the scene at the point of failure, the multimodal LLM infers if the subgoal's target object is present in the scene and if the action of the failed subgoal (Pickup, Desklamp) is feasible (Q2 and A3).
Based on this information, it decides whether to re-plan or the same subgoal needs to be attempted again due to uncertainty from the controller.
If a re-plan is necessary, the multimodal LLM explicitly infers the cause of failure, such as \textit{``The desk lamp is too heavy''}, and how to modify the plan based on vision information (Q4 and A6). 
Using this conversation, the Task Planner adjusts its initial plan to directly turn on the light without lifting the lamp.

\section{Conclusion}

This paper explores an effective zero-shot reasoning method for Embodied Instruction Following (EIF), the Socratic Planner. Leveraging Socratic self-QA-based task decomposition and visually grounded re-planning strategy, the Socratic Planner shows robust and outstanding performance in both offline and online settings. 
These contributions represent a significant advancement in enabling more effective task planning for embodied agents, showing the potential for following instructions in complex environments without relying on extensive labeled data.
\clearpage

\section*{Acknowledgements}
This work was partly supported by the IITP (RS-2021-II212068-AIHub/10\%, RS-2021-II211343-GSAI/15\%, 2022-0-00951-LBA/15\%, 2022-0-00953-PICA/20\%), NRF (RS-2024-00353991/20\%, RS-2023-00274280/10\%), and KEIT (RS-2024-00423940/10\%) grant funded by the Korean government.

\bibliographystyle{IEEEtran}
\bibliography{ref}

\end{document}